\documentclass[letterpaper]{article} 
\usepackage{aaai2026}  
\usepackage{times}  
\usepackage{helvet}  
\usepackage{courier}  
\usepackage[hyphens]{url}  
\usepackage{graphicx} 
\usepackage{natbib}  
\usepackage{caption} 
\usepackage{algorithm}
\usepackage{algorithmic}
\usepackage{multirow}
\usepackage{xcolor}
\usepackage{tabularx}
\usepackage{subcaption}
\usepackage{makecell}
\usepackage{booktabs}
\usepackage{amsmath, amssymb}
\usepackage{amsthm}
\newtheorem{definition}{Definition}
\usepackage{bbding}
\usepackage{newfloat}
\usepackage{listings}
\DeclareCaptionStyle{ruled}{labelfont=normalfont,labelsep=colon,strut=off} 
\floatstyle{ruled}
\newfloat{listing}{tb}{lst}{}
\floatname{listing}{Listing}
\newcommand{\stitle}[1]{\vspace*{0.15em}\noindent{\bf #1\/}}
\title{Region-Point Joint Representation for Effective Trajectory Similarity Learning}
\author{
    Hao Long\equalcontrib,
    Silin Zhou\equalcontrib,
    Lisi Chen,
    Shuo Shang\thanks{Shuo Shang is the corresponding author.}
}
\affiliations {
    University of Electronic Science and Technology of China\\
    haolong@std.uestc.edu.cn,
    \{zhousilinxy,jedi.shang\}@gmail.com,
    lchen012@e.ntu.edu.sg
}
\begin{document}
\maketitle

\begin{abstract}
Recent learning-based methods have reduced the computational complexity of traditional trajectory similarity computation, but state-of-the-art (SOTA) methods still fail to leverage the comprehensive spectrum of trajectory information for similarity modeling. To tackle this problem, we propose \textbf{RePo}, a novel method that jointly encodes \textbf{Re}gion-wise and \textbf{Po}int-wise features to capture both spatial context and fine-grained moving patterns. For region-wise representation, the GPS trajectories are first mapped to grid sequences, and spatial context are captured by structural features and semantic context enriched by visual features. For point-wise representation, three lightweight expert networks extract local, correlation, and continuous movement patterns from dense GPS sequences. Then, a router network adaptively fuses the learned point-wise features, which are subsequently combined with region-wise features using cross-attention to produce the final trajectory embedding. To train RePo, we adopt a contrastive loss with hard negative samples to provide similarity ranking supervision. Experiment results show that RePo achieves an average accuracy improvement of 22.2\% over SOTA baselines across all evaluation metrics.
\end{abstract}

\begin{links}
    \link{Code}{https://github.com/Alfred4c/RePo}
\end{links}

\section{Introduction}
The rapid proliferation of GPS-enabled devices and location-based services have led to the generation of massive amounts of trajectory data~\shortcite{bigdata,blue}. Trajectory similarity computation~\shortcite{similarity} is a fundamental task in trajectory analysis, supporting a wide range of downstream applications, such as trajectory classification~\shortcite{red}, urban planning~\shortcite{urbanplanning}, and anomaly detection~\shortcite{anomaly-detection2,anomalydetection3}.

\begin{figure}[!t]
    \centering
    \includegraphics[width=0.95\columnwidth]{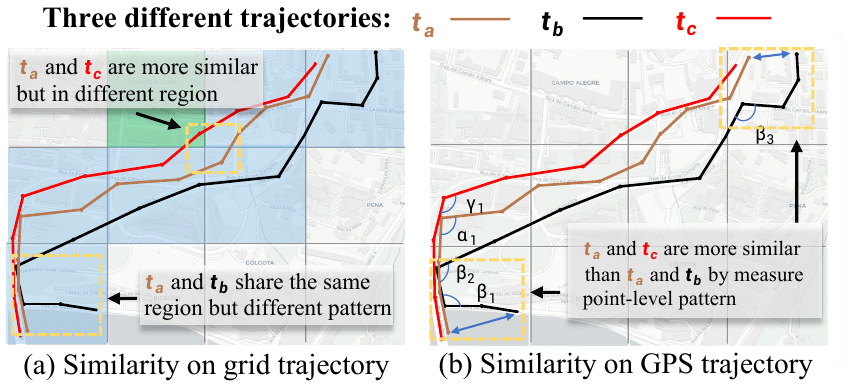}
    \caption{Comparison of trajectory similarity under grid (region) trajectory and GPS (point) trajectory.}
    \label{fig:figure1}
\end{figure}

Many heuristic distance measures, such as Discrete Fréchet (DFD)~\shortcite{DFD} and Dynamic Time Warping (DTW)~\shortcite{dtw}, compute trajectory similarity via exhaustive pairwise comparisons, resulting in poor scalability to large datasets~\shortcite{class-largedataset,class-largedataset2}. To address this issue, trajectory similarity learning methods~\shortcite{t2vec} aims to encode each trajectory into a vector representation, enabling efficient linear-time similarity computation via vector operations.

Existing trajectory similarity learning methods first transform GPS trajectories into grid trajectories to reduce irregularity in free space, and then learn trajectory embeddings from the resulting grid sequences~\shortcite{GREEN}. These approaches can be categorized into three types:
RNN-based models, e.g., t2vec~\shortcite{t2vec}, CL-Tsim~\shortcite{CLTsim}, leverage recurrent neural networks or contrastive learning to capture sequential dependencies;
Graph-based models, e.g., KGTS~\shortcite{KGTS}, TrajGAT~\shortcite{TrajGAT}, construct graphs to model spatial relationships or global dependencies among trajectory points;
Transformer-based models, e.g., SIMformer~\shortcite{SIMformer}, TrajCL~\shortcite{TrajCL}, employ self-attention mechanisms for efficient long-range dependency modeling.

While grid-based trajectories help mitigate GPS irregularities in free space and effectively preserve the main structure of trajectories, they inevitably compromise the descriptive richness of trajectory representations. As illustrated in Figure~\ref{fig:figure1}(a), mapping trajectories $t_a$, $t_b$, and $t_c$ onto grid sequences smooths out local noise and highlights their global spatial structure; for instance, $t_a$ and $t_b$ become identical at the region level after grid mapping. This grid abstraction, however, introduces two key limitations:
1) Although $t_a$ and $t_b$ become identical after grid mapping, $t_a$ is actually more similar to $t_c$, even though $t_c$ follows a different grid sequence.
2) Intra-grid variations, such as local deviations and sharp turns, are lost, making it difficult to capture fine-grained movement details. In essence, although grid representations are robust to irregularity and highlight the overall spatial structure, they may obscure subtle characteristics and fail to capture detailed motion dynamics.

Given these limitations, it is instructive to consider the intrinsic information captured at the point-level. As shown in Figure~\ref{fig:figure1}(b), although three trajectories fall within the same grid cell (highlighted in red), point-level information enables a clear distinction among them. Specifically, point-wise features encode three key aspects: (1) \textit{variability}, capturing dynamic changes in inter-point distances or turning angles; (2) \textit{locality}, preserving fine-grained geometric and spatial patterns; and (3) \textit{continuity}, maintaining the smooth progression of movement. These motivate us to combine the robustness of grid-based representations with point-level information for more discriminative trajectory representation.

In this light, we propose \textbf{RePo}, an efficient multimodal trajectory similarity learning method that jointly learns trajectory embeddings from both region-wise and point-wise perspectives. For region-wise learning, we consider two modalities to introduce inter-grid information, i.e., structural context, and intra-grid information, i.e., visual context. Structural context embedding captures spatial dependencies and transition patterns between regions by constructing a grid-level transition graph and learning global inter-region correlations. Visual context embedding captures intra-grid semantics by extracting environmental features from the satellite image of each grid cell, enriching the grid representation with detailed local visual cues. For point-wise learning, we employ locality-aware, correlation-aware, and continuity-preserving encoding modules to comprehensively capture local details, dynamic correlations, and continuity at the point level. Region-wise and point-wise features are fused with cross-attention, and the model is trained by supervised contrastive loss.

We conduct extensive experiments on real-world trajectory datasets and compare with seven SOTA baselines under three widely used trajectory similarity measures, i.e., DTW, DFD, and EDwP~\shortcite{edwp}. Experiment results show that RePo consistently outperforms all baselines across different datasets and distance measures, with an average accuracy improvement of 22.2\%. Ablation studies confirm the effectiveness of RePo’s designs. Generalization and visualized results show that RePo can effectively capture true trajectory similarity and generalize well to large-scale datasets.

Our contributions are summarized as follows:
\begin{itemize}
    \item We observe that existing SOTA methods, which rely solely on grid-based representations, fail to capture fine-grained variations in trajectory details for similarity.

    \item We propose RePo, a novel framework that jointly learns region-wise and point-wise representations to construct unified trajectory embeddings with a strong capacity to capture fine-grained and discriminative spatial patterns for similarity measurement.

    \item We evaluate RePo by comparing it against seven state-of-the-art methods across multiple popular similarity measures, conducting comprehensive ablation studies, and evaluating its generalization performance.
\end{itemize}

\section{Related Work}
\stitle{Heuristic-based Similarity Measures.} 
Traditional trajectory similarity measures can be classified into two categories. \emph{Point-based methods}, such as Euclidean Distance, Dynamic Time Warping (DTW)~\cite{dtw}, LCSS~\cite{lcss}, and EDR~\cite{edr}, compare individual points along the trajectories to compute similarity. \emph{Shape-based methods}, like Hausdorff Distance~\cite{hausdorff} and Fréchet Distance~\cite{frechet}, focus on the overall geometric shape, measuring maximum deviation between curves. While these methods provide interpretable similarity scores, they generally require explicit pairwise matching, resulting in $\mathcal{O}(l^2)$ computational complexity, where $l$ is the trajectory length, thus limiting scalability to large datasets.

\stitle{Learning-based Similarity Measures.}
Recent deep learning approaches address the limitations of heuristic methods by learning trajectory-level embeddings for efficient similarity computation. RNN-based models, such as t2vec~\cite{t2vec}, NeuTraj~\cite{NeuTraj}, and CL-Tsim~\cite{CLTsim}, employ recurrent networks to capture sequential and spatial-temporal dependencies. Graph-based models (e.g., TrajGAT~\cite{TrajGAT}, GTS~\cite{GTS}, KGTS~\cite{KGTS}) leverage graph structures and attention mechanisms to model spatial relationships among trajectory points. Transformer-based models, such as SIMformer~\cite{SIMformer} and TrajCL~\cite{TrajCL}, utilize self-attention for long-range dependency modeling and fine-grained feature extraction. Nevertheless, most existing methods fail to fully exploit the rich information present in trajectories. In contrast, our approach integrates region-wise and point-wise in a unified framework for comprehensive trajectory similarity learning.

\section{Preliminaries}

\subsection{Definition}
\begin{definition}[GPS Trajectory] A GPS trajectory $T_{gps}$ is defined as a sequence of points collected at a fixed sampling time interval, where each point $p_{i} = (lon_{i}, lat_{i})$ of $T_{gps}$ represents the $i$-th sampled location, consisting of a longitude $lon_{i}$ and a latitude $lat_{i}$.
\end{definition}

\begin{definition}[Grid Trajectory]
A grid trajectory $T_{grid}$ is a sequence of discrete grid cells over the city map. Each element $g_i \in T_{grid}  = \left< g_1, g_2, ..., g_{n_1} \right>$ represents a grid cell ID, obtained by mapping GPS points of $T_{gps}$ to the grid space.
\end{definition}

\subsection{Problem Statement}
\stitle{Trajectory Similarity Learning.}
Given a similarity measure \(g(\cdot, \cdot)\) (e.g., DTW, DFD), trajectory similarity learning aims to learn a function \(f(\cdot, \cdot)\) that approximates \(g(\cdot, \cdot)\) for any pair of trajectories \((T_i, T_j)\) in a \(d\)-dimensional embedding space, i.e., \( f(T_i, T_j) \approx g(T_i, T_j) \). The function \(f(\cdot, \cdot)\) is trained to preserve the relative similarity ranking of trajectories. That is, for any three different trajectories \(T_i, T_j, T_k\), if \(g(T_i, T_j) < g(T_i, T_k)\), then it should follow that \(f(T_i, T_j) < f(T_i, T_k)\).

\section{The RePo Method}
\begin{figure}[!t]
\centering
\includegraphics[width=\columnwidth]{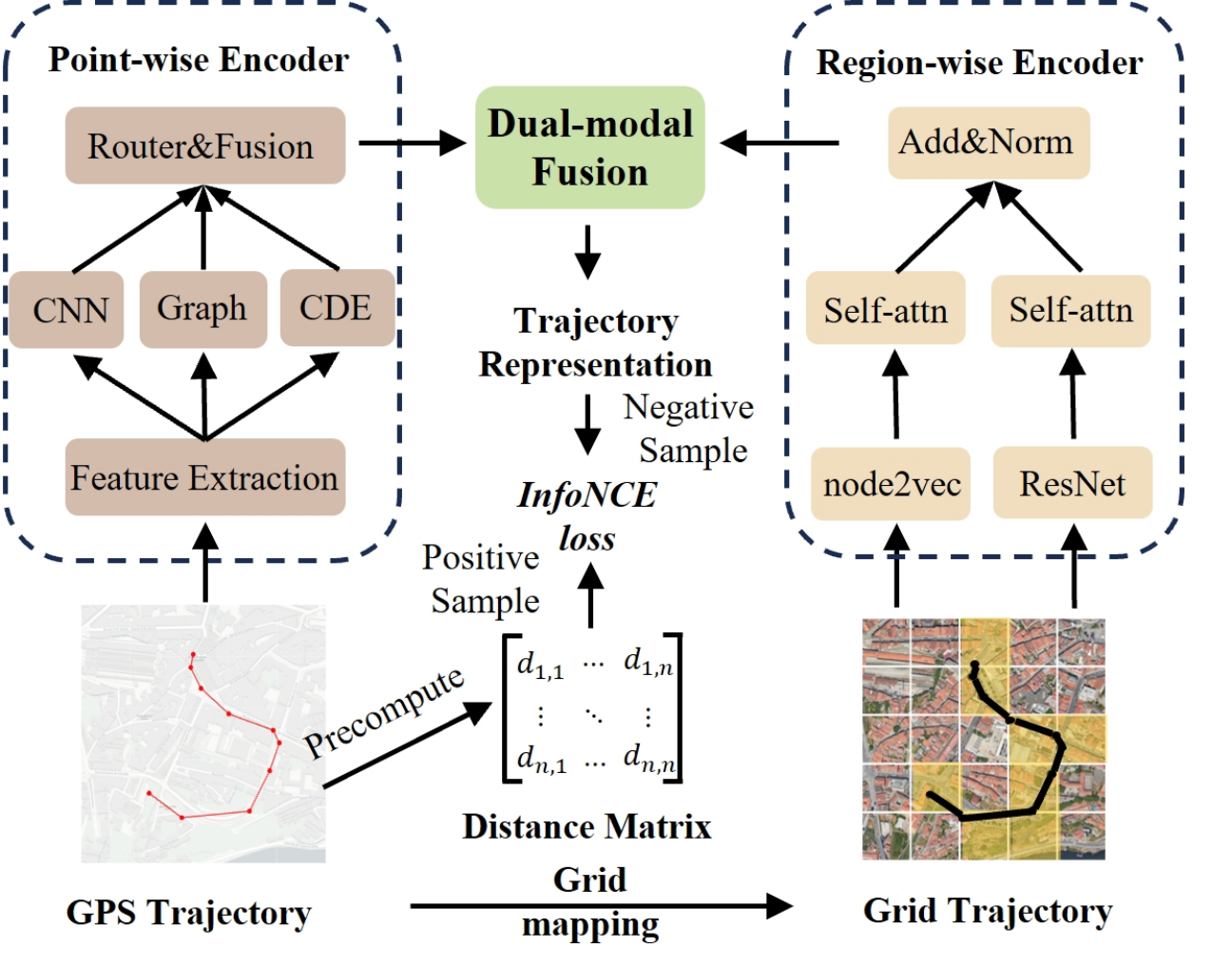}
\caption{An overview of our RePo Method}
\label{fig2}
\end{figure}
Figure~\ref{fig2} shows the overall architecture of RePo. The model consists of three key components: 1) \textit{Region-wise Encoder}: it learns regional embeddings from the grid-based trajectory by incorporating both spatial structural features and visual contextual information; 2) \textit{Point-wise Encoder}: it captures point embeddings from the GPS trajectory by modeling three types of encodings: locality, correlation, and continuity; 3) \textit{Dual-modal Fusion}: it integrates point and grid embedding to generate the trajectory embedding. We train our model with a supervised contrastive loss, using positive and hard negative samples based on trajectory distances to learn similarity rankings. We use \texttt{[CLS]} token to learn trajectory embedding. For simplicity, multi-head attention and the \texttt{[CLS]} token are omitted in the formulas.

\subsection{Region-wise Encoder}
The region-wise encoder captures global trajectory semantics at the grid level through two complementary representations: structural context embedding, which models inter-region spatial correlations, and visual context embedding, which provides visual features for each region.

\stitle{Structural Context Embedding.} Trajectory movement is inherently constrained by the environment. For example, a vehicle can only move through grid cells that overlap with road segments. To capture the spatial dependencies and transition patterns between different regions, we construct a transition graph $\mathcal{G} \in \mathbb{R}^{M \times M}$ from the training trajectory dataset, where $M$ denotes the total number of grid cells. An edge $\mathcal{G}[i][j] = 1$ is established if there exists a trajectory that moves from grid cell $g_i$ to $g_j$. We do not incorporate transition frequencies into the graph, as they are often imbalanced and less indicative of the underlying road structure context. Then we use Node2vec~\cite{node2vec} to learn global spatial correlation grid embeddings as follows:
\begin{equation}
    \mathbf{E}^s = \mathrm{Node2Vec}(\mathcal{G}),
\end{equation}
where $\mathbf{E}^s \in \mathbb{R}^{M \times d}$. Then we can transform a grid trajectory sequence $T_{grid} = \left< g_1, g_2, ..., g_{n_1} \right>$ into a structural context embedding sequence $\mathbf{E}^s = \left< \mathbf{e}_1^s, \mathbf{e}_2^s, ..., \mathbf{e}_{n_1}^s \right>$ by looking up the embedding table $\mathcal{E}^s$. Finally, we add position encoding~\cite{transformer} and use a self-attention to learn sequence-level correlations by following:
\begin{equation}
    \mathbf{H}^s = \mathrm{SelfAttn}(\mathbf{E}^s) \in \mathbb{R}^{n_1 \times d}.
\end{equation}

\stitle{Visual Context Embedding.} In addition to spatial correlations, the visual semantics of grid cells provide rich contextual information. For example, trajectories tend to avoid areas dominated by buildings and are more likely to traverse regions with roads. Easily accessible satellite imagery corresponding to grid cells (e.g., from Google Maps) can reveal such features. To capture these visual semantics, we use ResNet~\cite{resnet} to process the satellite images corresponding to each grid cell and extract visual representations. Given the city map $\mathcal{I} \in \mathbb{R}^{H \times W}$, where $H \times W = M$ and each element in $\mathcal{I}$ corresponds to the satellite image of a grid cell, we extract visual context embeddings as follows:
\begin{equation}
    \mathbf{E}^v = \mathrm{ResNet}(\mathcal{I}),
\end{equation}
where $\mathbf{E}^v \in \mathbb{R}^{M \times d}$. Similar to structural context embedding, we apply position encoding and apply self-attention to learn sequence-level correlations:
\begin{equation}
    \mathbf{H}^v = \mathrm{SelfAttn}(\mathbf{E}^v)\in \mathbb{R}^{n_1 \times d},
\end{equation}
Therefore, the final trajectory embedding on region context information can be obtained as follows:
\begin{equation}
    \mathbf{H}^r = \mathbf{H}^s + \mathbf{H}^v.
\end{equation}

 \subsection{Point-wise Encoder}

The point-wise encoder learns trajectory representations through three expert modules: locality-aware encoding for local details, correlation-aware encoding for variability, and continuity-preserving encoding for smooth transitions.

\stitle{Feature Extraction.} 
Since GPS data contains only longitude and latitude, directly inputting it provides limited spatial context. To enhance trajectory representation, we first extract spatial features for each point by converting geographic coordinates to Mercator projection coordinates $(x_i, y_i)$~\cite{TrajCL}, thus mitigating geographic scale effects. Then we compute the distance and bearing angle between $p_i$ and its adjacent points $p_{i-1}$ and $p_{i+1}$. The resulting feature for each trajectory point is a vector defined as:
\begin{equation}
\mathbf{s}_i = \left( x_i, y_i, d_{i-1,i}, \theta_{i-1,i}, d_{i,i+1}, \theta_{i,i+1} \right),
\end{equation}
where $(x_i, y_i)$ denote the normalized Mercator coordinates, $d_{i-1,i}$ and $d_{i,i+1}$ represent the distances between $p_i$ and its previous and next points, respectively, and $\theta_{i-1,i}$, $\theta_{i,i+1}$ are the corresponding bearing angles. Then we use a Linear layer to learn point embedding as follows:
\begin{equation}
    \mathbf{e}_i^p = \mathrm{Linear}(\mathbf{s}_i) \in \mathbb{R}^{d},
\end{equation}
therefore, a $n_2$-length GPS trajectory $T_{gps}$ can be transformed into a feature embedding $\mathbf{E}^{p} \in \mathbb{R}^{n_2 \times d}$. 
 \begin{figure}[!t]
     \centering
     \includegraphics[width=\columnwidth]{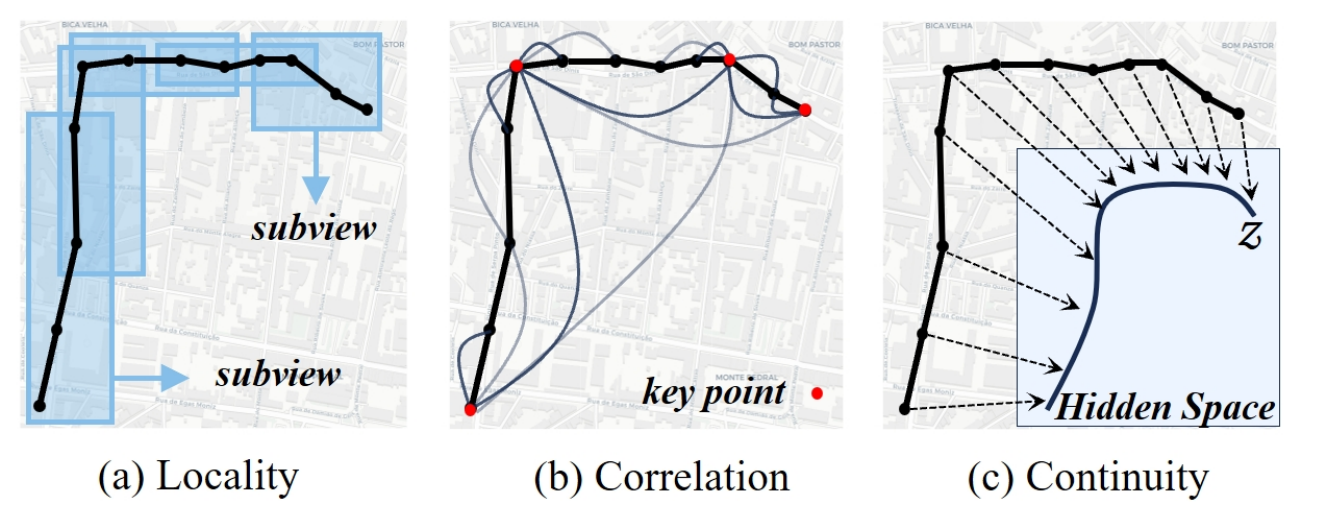}
     \caption{Three types of encodings in Point-wise encoder.}
     \label{fig3}
 \end{figure}

\stitle{Locality-aware Encoding.} 
Locality refers to the strong short-range dependency between adjacent points in a trajectory, reflecting the fine-grained geometric and dynamic patterns over short segments. In a trajectory, each point is highly correlated with its immediate neighbors, capturing local continuity and motion trends. As shown in Figure~\ref{fig3}(a), we extract overlapping local sub-segments along the trajectory to focus on these local details, enabling the model to effectively capture and encode fine-grained movement patterns. To effectively capture such local dependencies, we employ a multi-layer 1D Convolutional Neural Network (CNN)~\cite{cnn}, which is well-suited for extracting fine-grained locality information by modeling short-range spatial patterns along the trajectory. This ensures that local dynamics are adequately preserved. The locality-aware encoding is learned as follows:
\begin{equation}
    \mathbf{H}^{loc}_{(l)} = \text{LeakyReLU}(\text{GN}(\text{Conv}(\mathbf{H}^{loc}_{(l-1)})))  \in \mathbb{R}^{n_2 \times d},
\end{equation}
where $l$ is the layer number and $\mathbf{H}^{loc}_{(0)} = \mathbf{E}^{p}$ is the initial embedding of the trajectory. The final output $\mathbf{E}^{loc}=\mathbf{H}^{loc}_{(l)}$ encodes the local movement patterns within a trajectory.

\stitle{Correlation-aware Encoding.} 
Global correlation refers to the semantic relationships and dependencies between non-adjacent points in a trajectory, which are essential for capturing overall trajectory variability and complex movement patterns. We represent each trajectory as an adaptive graph, where edges are learned based on the semantic similarity between points. This enables the model to capture both strong and weak dependencies across the entire sequence. As illustrated in Figure~\ref{fig3}(b), the connections between key points visualize how global correlations of varying strength are modeled within the trajectory.

Given the trajectory feature matrix $\mathbf{E}^{p} \in \mathbb{R}^{n_2 \times d}$, node embeddings are obtained via a Linear projection:
\begin{equation}
    \mathbf{E} = \text{Linear}(\mathbf{E}^{p}) \in \mathbb{R}^{n_2 \times d},
\end{equation}
Next, an adjacency matrix $\mathcal{A} \in \mathbb{R}^{n_2 \times n_2}$ for a trajectory is constructed based on node similarity:
\begin{equation}
    \mathcal{A} = \mathrm{Softmax}\left( \mathrm{ReLU}(\mathbf{E} \mathbf{E}^\top) \right).
\end{equation}
Structural information is propagated and normalized as:
\begin{equation}
    \mathbf{E}^{cor} = \mathrm{LayerNorm}(\mathcal{A}\mathbf{E}W),
\end{equation}
where $W \in \mathbb{R}^{d \times d}$ is a learnable matrix.

\stitle{Continuity-preserving Encoding.} 
Continuity at the representation space is essential for faithfully preserving the smooth transitions in trajectories, ensuring that the learned embeddings change gradually along the path. To achieve this, we use a neural controlled differential equation (CDE) encoder~\cite{cde}, which interpolates discrete feature sequences into a continuous path in the representation space. As illustrated in Figure~\ref{fig3}(c), this approach enables the model to capture consistent and smooth transitions in the hidden space, providing a natural encoding of trajectory continuity.

Given $\mathbf{E}^{p} \in \mathbb{R}^{n_2 \times d}$, we interpolate the feature sequence using Hermite cubic splines~\cite{cubic} to obtain a continuous path. The initial hidden state $\mathbf{z}_0$ is computed from the first point in the trajectory, where $e$ denotes the feature vector of the first point. Then, the initial hidden state is computed as:
\begin{equation}
    \mathbf{z}_0 = \text{MLP}(e) = \phi_2(\phi_1(e)),
\end{equation}
where $\phi_1$ and $\phi_2$ are linear layers with ReLU activation. The evolution of the hidden state is governed by the neural CDE:
\begin{equation}
    \mathbf{z}(t_\ell) = \mathbf{z}_0 + \int_{t_0}^{t_\ell} f_\theta(\mathbf{z}(s)) \, d\mathbf{X}(s),
\end{equation}
where $f_\theta(\mathbf{z}(t))$ is a neural network function representing the rate of change of the hidden state. The final continuity-preserving embedding can be obtained by:
\begin{equation}
    \mathbf{E}^{con} = \text{Linear}(\mathbf{z}) \in \mathbb{R}^{n_2 \times d}.
\end{equation}

\stitle{Mixture-of-Experts Dynamic Fusion.}
Since the three encoding modules capture different aspects of trajectory information, we design a dynamic fusion mechanism to adaptively integrate locality, correlation, and continuity cues. For each position in the trajectory, the importance of each expert is dynamically determined based on the local features at that position, allowing the model to adjust the fusion strategy according to the content of each part of the trajectory.

We adopt a mixture-of-experts (MoE) paradigm~\cite{moe,moe2}, where the outputs of the three trajectory encoding modules serve as distinct experts. A shared MLP network (router) predicts the importance weights of each expert, enabling adaptive, position-specific fusion. The final fused feature at each position is computed as a weighted sum:
{\small
\begin{equation}
    \mathbf{w} = \mathrm{Softmax}\big(
        \mathrm{MLP}(\mathbf{E}^{loc}),~
        \mathrm{MLP}(\mathbf{E}^{cor}),~
        \mathrm{MLP}(\mathbf{E}^{con})
    \big),
\end{equation}
}where $\mathbf{w} \in \mathbb{R}^{n_2 \times 3}$ denotes the importance scores for all sequence positions. Let $w^{(i)}_{j}$ denote the $i$-th element of the weight vector at position $j (i = 1, 2, 3)$, corresponding to the locality, correlation, and continuity experts, respectively. The fused feature at each position is computed as:
\begin{equation}
    \mathbf{H}^{p}_{j} = w^{(1)}_{j} \mathbf{E}^{loc}_{j} + w^{(2)}_{j} \mathbf{E}^{cor}_{j} + w^{(3)}_{j} \mathbf{E}^{con}_{j}.
\end{equation}
This position-wise dynamic fusion enables the model to flexibly leverage the complementary strengths of different experts throughout the trajectory.

\subsection{Dual-modal Fusion}
The dual-modal fusion module integrates point and grid embeddings to generate the final trajectory representation. We use cross-attention with point embeddings as queries and grid embeddings as keys and values, because each trajectory point benefits from actively attending to global spatial context. This setup enables each point to selectively aggregate relevant information from the entire grid, allowing the model to enrich local features with comprehensive global cues and produce a more expressive representation.

To seamlessly integrate region-wise and point-wise trajectory information, we design a cross-modal attention fusion module. Specifically, the grid representation $\mathbf{H}^r \in \mathbb{R}^{n_1 \times d}$ serves as the query $\mathbf{Q}$, while the point representation $\mathbf{H}^p \in \mathbb{R}^{n_2 \times d}$ serves as the key $\mathbf{K}$ and value $\mathbf{V}$. The cross-attention operation is formulated as:
\begin{equation}
    \mathbf{H}^o = \mathrm{CrossAttn}(\mathrm{q}=\mathbf{H}^r,\mathrm{k}=\mathbf{H}^p,\mathrm{v}=\mathbf{H}^p),
\end{equation}
where appropriate padding masks are applied for variable-length sequences.

The cross-attended features $\mathbf{H}^o$ are combined with the query via residual connection and layer normalization, and then passed through a feed-forward network (FFN) to yield the final output of the fusion layer:
\begin{equation}
    \mathbf{E} = \mathrm{FFN}(\mathrm{LayerNorm}(\mathbf{H}^o + \mathbf{H}^r)) + \mathbf{H}^o,
\end{equation}
where $\mathbf{E} \in \mathbb{R}^{n_1 \times d}$. Finally, we use the \texttt{[CLS]} token of $\mathbf{E}$ as the final trajectory embedding.

\subsection{Supervised Contrastive Loss}
To train RePo, we employ a supervised InfoNCE-based~\cite{infoce} loss that leverages precomputed trajectory distance matrices. Specifically, we compute a pairwise ground-truth distance matrix $\mathcal{D}$ using trajectory distance measures. For each anchor trajectory $T_i$, the positive sample $j^+$ is selected as its nearest neighbor in $\mathcal{D}$, while hard negatives are selected from the batch based on cosine similarity in the embedding space (excluding the anchor and positive). For each pair of trajectories $T_i$ and $T_j$, the similarity $S_{i,j}$ is computed as the cosine similarity between their final embeddings. The loss is defined as:
\begin{equation}
\small
    \mathcal{L} = - \frac{1}{N} \sum_{i=1}^N \left[
        \frac{ S_{i,j^+} }{\tau }
        - \log \sum_{j \in \mathcal{N}_i^-} \exp\left( \frac{ S_{i,j}}{\tau} \right)
    \right],
\end{equation}
where $\tau$ is a temperature hyperparameter and $\mathcal{N}_i^-$ denotes the set of $k$ hardest negatives for anchor $i$.

We select positives using ground-truth trajectory distances to ensure true semantic similarity, while negatives are mined in the representation space to encourage the model to distinguish hard cases under the current embedding. This design leverages both accurate distance supervision and adaptive discrimination, promoting a more robust and semantically meaningful embedding space.

\subsection{Complexity Analysis}
The overall computational complexity of RePo is dominated by the attention mechanisms within each encoder block. For a trajectory of length $L$ and hidden dimension $d$, both the self-attention and cross-attention modules require $\mathcal{O}(L^2 d)$ operations due to pairwise sequence interactions. In addition, the convolutional, graph-based, and neural CDE modules, as well as the feed-forward sublayers, each require $\mathcal{O}(L d^2)$ operations. Therefore, the total per-layer complexity is $\mathcal{O}(L^2 d + L d^2)$,  where the quadratic attention term $\mathcal{O}(L^2 d)$ dominates for longer sequences.

\section{Experimental Evaluation}

\subsection{Experiment Settings}
\stitle{Datasets.} We conduct experiments on three widely used real-world trajectory datasets:
\textbf{Porto}~\cite{porto}, \textbf{Geolife}~\cite{geolife} and \textbf{Chengdu}. The points of trajectories in Porto, Chengdu, and Geolife are sampled every 15 seconds, 30 seconds, and 5 seconds. For all datasets, we retain trajectories with lengths between 10 and 300 points. Following previous studies~\cite{NeuTraj, TrajGAT}, we sampled 10,000 trajectories from each dataset and randomly split them into 20\% training, 10\% validation, and 70\% testing sets.

\stitle{Baselines.} We compare RePo with 7 representative trajectory similarity learning methods as follows: 
\begin{itemize}
    \item \textbf{t2vec}~\cite{t2vec}: it learns trajectory embeddings using RNN and a spatial proximity-aware loss.
    \item 
    \textbf{NeuTraj}~\cite{NeuTraj}: it is based on LSTM with spatial memory for supervised similarity learning.
    \item 
    \textbf{CL-TSim}~\cite{CLTsim}: it employs contrastive learning and a Siamese network for trajectory similarity search.
    \item \textbf{TrajGAT}~\cite{TrajGAT}: it models each trajectory as a hierarchical graph and applies graph attention networks.
    \item
    \textbf{TrajCL}~\cite{TrajCL}: it adopts self-supervised contrastive learning with dual-feature self-attention. 
    \item 
    \textbf{KGTS}~\cite{KGTS}: it incorporates spatial and semantic knowledge into trajectory representation.
    \item \textbf{SIMformer}~\cite{SIMformer}: it leverages a transformer encoder for efficient similarity approximation.

\end{itemize}

\setlength{\tabcolsep}{1.5mm} 
\begin{table*}[ht]
\centering
\small
\scalebox{0.9}{
\begin{tabular}{l|l|cccc|cccc|cccc}
\toprule
\multirow{2}{*}{Dataset} 
& \multirow{2}{*}{Method}
& \multicolumn{4}{c|}{DFD}
& \multicolumn{4}{c|}{DTW}
& \multicolumn{4}{c}{EDwP} \\
\cmidrule{3-6} \cmidrule{7-10} \cmidrule{11-14}
&
& HR@1 & R5@20 & MRR & NDCG
& HR@1 & R5@20 & MRR & NDCG
& HR@1 & R5@20 & MRR & NDCG \\
\midrule
\multirow{9}{*}{Porto} 
&t2vec      & 0.213 & 0.556 & 0.337 & 0.433 & 0.239 & 0.624 & 0.373 & 0.490 & 0.235 & 0.612 & 0.366 & 0.494 \\
&CL-TSim    & 0.101 & 0.317 & 0.184 & 0.213 & 0.116 & 0.362 & 0.204 & 0.266 & 0.114 & 0.357 & 0.203 & 0.273 \\
&NeuTraj    & 0.372 & \underline{0.840} &0.536 & \underline{0.755}& \underline{0.308} & \underline{0.807} & \underline{0.448} & \underline{0.672} & \underline{0.346} &\underline{0.787}   & \underline{0.501} &\underline{0.647}  \\
&TrajCL     & 0.326 & 0.803 & 0.483 & 0.687 & 0.217 & 0.603 & 0.343 & 0.556 & 0.197 & 0.586 & 0.328 & 0.493 \\
&TrajGAT    & 0.346 & 0.748 & 0.497 & 0.475 & 0.296 & 0.683 & 0.437 & 0.450 & 0.161 & 0.413 & 0.261 & 0.298 \\
&KGTS       & 0.215 & 0.392 & 0.321 & 0.333 & 0.253 & 0.444 & 0.369 & 0.411 & 0.234 & 0.420 & 0.345 & 0.388 \\
&SIMformer      & \underline{0.422} & 0.828 & \underline{0.575} & 0.714 & 0.300 & 0.755 & 0.446 & 0.651 & 0.321 & 0.579 & 0.466 & 0.475 \\
&\textbf{RePo}   & \textbf{0.483} & \textbf{0.922} & \textbf{0.640} & \textbf{0.771} & \textbf{0.532} & \textbf{0.955} & \textbf{0.682} & \textbf{0.821} & \textbf{0.512} & \textbf{0.947} & \textbf{0.667} & \textbf{0.833} \\
\cmidrule{2-14}
&\textbf{Improv.}       &\textbf{14.5\%} &\textbf{9.76\%}  & \textbf{11.3\%} &\textbf{2.11\%}  & \textbf{72.7\%} & \textbf{18.3\%} & \textbf{52.2\%} & \textbf{22.1\%} & \textbf{47.9\%} & \textbf{20.3\%} & \textbf{33.1\%} & \textbf{28.7\%} \\
\bottomrule
\midrule
\multirow{9}{*}{Geolife}
&t2vec      & 0.180 & 0.401 & 0.260 & 0.334 & 0.188 & 0.430 & 0.273 & 0.368 & 0.171 & 0.429 & 0.256 & 0.368 \\
&CL-TSim    & 0.166 & 0.506 & 0.270 & 0.430 & 0.181 & 0.571 & 0.295 & \underline{0.497} & 0.173 & 0.556 & 0.285 & 0.500 \\
&NeuTraj    & 0.289 & 0.713 & 0.403 & 0.630 & 0.202 & 0.532 & 0.287 & 0.455 & \underline{0.250} & \underline{0.657} & \underline{0.363} & \underline{0.581} \\
&TrajCL     & 0.129 & 0.479 & 0.238 & 0.457 & 0.107 & 0.357 & 0.195 & 0.361 & 0.062 & 0.285 & 0.134 & 0.316 \\
&TrajGAT    & 0.283 & 0.719 & 0.426 & 0.574 & 0.212 & \underline{0.587} & \underline{0.335} & 0.477 & 0.214 & 0.569 & 0.326 & 0.501 \\
&KGTS       & 0.206 & 0.308 & 0.283 & 0.338 & \underline{0.219} & 0.324 & 0.296 & 0.366 & 0.210 & 0.323 & 0.289 & 0.369 \\
&SIMformer      & \underline{0.316} & \underline{0.730} & \underline{0.452} & \underline{0.677} & 0.194 & 0.525 & 0.288 & 0.474 & 0.218 & 0.570 & 0.342 & 0.560 \\
&\textbf{RePo}   & \textbf{0.325} & \textbf{0.745} & \textbf{0.455} & \textbf{0.678} & \textbf{0.342} & \textbf{0.776} & \textbf{0.475} & \textbf{0.685} & \textbf{0.347} & \textbf{0.787} & \textbf{0.483} & \textbf{0.728} \\
\cmidrule{2-14}
&\textbf{Improv.}       &\textbf{2.84\%} &\textbf{2.05\%}  & \textbf{0.66\%} &\textbf{0.14\%}  & \textbf{56.5\%} & \textbf{32.2\%} & \textbf{41.8\%} & \textbf{37.8\%} & \textbf{38.8\%} & \textbf{19.8\%} & \textbf{33.1\%} & \textbf{25.3\%} \\
\bottomrule
\midrule
\multirow{9}{*}{Chengdu}
&t2vec      & 0.306 & 0.737 & 0.462 & 0.533 & \underline{0.330} & \underline{0.792} & \underline{0.493} & 0.602 & \underline{0.324} & \underline{0.768} & \underline{0.487} & 0.596 \\
&CL-TSim    & 0.094 & 0.400 & 0.190 & 0.310 & 0.114 & 0.444 & 0.215 & 0.367 & 0.103 & 0.431 & 0.204 & 0.366 \\
&NeuTraj    & 0.316 & 0.830 & 0.478 & \underline{0.777} & 0.249 & 0.722 & 0.385 & \underline{0.609} & 0.263 & 0.740 & 0.409 & 0.657 \\
&TrajCL     & 0.313 & 0.859 & 0.482 & 0.768 & 0.110 & 0.409 & 0.206 & 0.354 & 0.091 & 0.407 & 0.184 & 0.376 \\
&TrajGAT    & 0.338 & 0.824 & 0.505 & 0.542 & 0.190 & 0.618 & 0.327 & 0.447 & 0.100 & 0.388 & 0.193 & 0.321 \\
&KGTS       & 0.179 & 0.420 & 0.294 & 0.403 & 0.225 & 0.491 & 0.353 & 0.485 & 0.186 & 0.432 & 0.297 & 0.463 \\
&SIMformer       & \underline{0.387} & \underline{0.862} & \underline{0.546} & 0.763 & 0.235 & 0.706 & 0.369 & 0.608 & 0.279 & 0.755 & 0.429 & \underline{0.667} \\
&\textbf{RePo}   & \textbf{0.392} & \textbf{0.895} & \textbf{0.552} & \textbf{0.796} & \textbf{0.428} & \textbf{0.929} & \textbf{0.592} & \textbf{0.823} & \textbf{0.360} & \textbf{0.904} & \textbf{0.526} & \textbf{0.830} \\
\cmidrule{2-14}
&\textbf{Improv.}       &\textbf{1.29\%} &\textbf{3.82\%}  & \textbf{1.09\%} &\textbf{2.44\%}  & \textbf{29.6\%} & \textbf{17.3\%} & \textbf{20.1\%} & \textbf{35.1\%} & \textbf{11.1\%} & \textbf{ 17.7\%} & \textbf{8.00\%} & \textbf{24.4\%} \\
\bottomrule
\end{tabular}
}
\caption{Performance comparison on the three datasets under different trajectory similarity metrics. Best results are bolded, second-best are underlined. “Improv” shows our improvement over the strongest baseline.
}
\label{tab:porto-results}
\end{table*}

\begin{figure*}[!t]
\centering
\includegraphics[width=\textwidth]{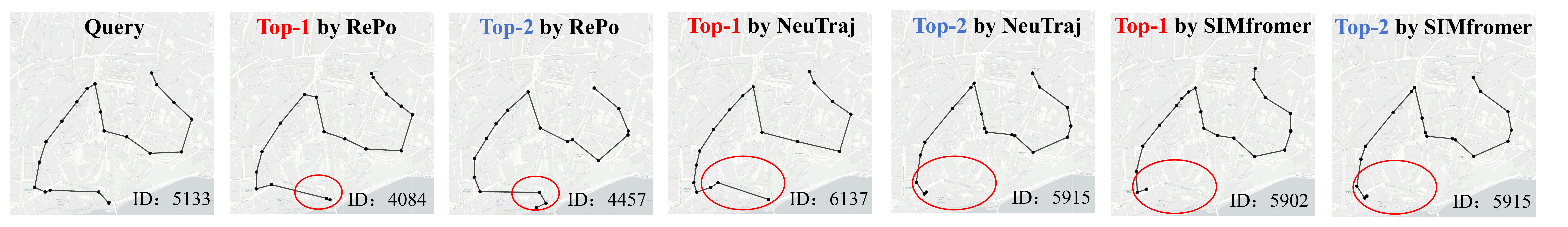}
\caption{
Trajectory visualization results on the Porto dataset. We compare the top-2 retrievals produced by RePo, NeuTraj, and SIMformer. Significant differences in visualization results are highlighted with red circles.
}

\label{fig:visual}
\end{figure*}

\stitle{Implementation Details.} All experiments are conducted on a single NVIDIA GeForce RTX 3090 GPU. The embedding dimension is set to 512. The spatial grid for mapping GPS points is constructed at zoom level 18, with each grid cell covering roughly 153\,m~$\times$~153\,m at the equator. The self-attention layer of the Region-wise encoder is set to 1. The CNN layer of the Point-wise encoder is set to 3. We use a temperature hyperparameter $\tau=0.2$ for the contrastive loss. The Adam optimizer is used with a learning rate of $2 \times 10^{-5}$.

\stitle{Evaluation Metrics.}
Following prior studies~\cite{TrajGAT, TrajCL}, we use top-$k$ Hit Ratio (HR@$k$) and top-$k$ Partial Hit Ratio (R$m$@$k$) for trajectory similarity retrieval and ranking evaluation. Besides, we further introduce two widely used ranking metrics, i.e., Mean Reciprocal Rank (MRR) and Normalized Discounted Cumulative Gain (NDCG), to provide a comprehensive evaluation. Higher values indicate better performance.

\subsection{Main Results}
We report results on the Porto, Geolife, and Chengdu datasets under three trajectory similarity measures (DFD, DTW, EDwP) and four evaluation metrics (HR@1, R5@20, MRR, NDCG@50) in Table~\ref{tab:porto-results}, more results are shown in the Appendix. RePo consistently achieves the best performance across all metrics and datasets, demonstrating strong robustness to different similarity measures and data distributions. Notably, RePo outperforms the best baselines by up to 72.7\% on HR@1 (DTW, Porto), and delivers stable gains in both retrieval and ranking metrics. 

RePo’s explicit fusion of region-wise and point-wise representations enables comprehensive trajectory modeling, resulting in higher retrieval accuracy and more robust ranking. While models such as SIMformer and NeuTraj achieve competitive results on certain metrics, their performance fluctuates across datasets and evaluation criteria. 

Performance on Geolife is generally lower for all methods, including RePo, likely due to its sparser trajectories and higher urban complexity, which make discrimination more difficult. Notably, improvements under the DFD metric are less pronounced compared to DTW and EDwP. This is because DFD focuses primarily on the global geometric shape and overall alignment of trajectories, while our model emphasizes the capture of multi-scale spatial patterns. In contrast, DTW and EDwP are more sensitive to local similarities and flexible alignments, allowing our discriminative multi-modal representations to demonstrate larger gains.

\stitle{Trajectory Retrieval Visualization.}
To qualitatively assess retrieval performance, we visualize the top-2 results from RePo, SIMformer, and NeuTraj for a random query (Figure~\ref{fig:visual}). RePo retrieves trajectories that better match both global and local patterns, highlighting its superior ability to capture comprehensive trajectory similarity.

\subsection{Micro Results}
\noindent\textbf{Ablation Study.}
We create several model variants by removing individual modules as follows:
\begin{itemize}
    \item w/o ResNet: removing the ResNet network.
    \item w/o node2vec: removing the node2vec network.
    \item w/o CNN: removing the locality-aware encoding.
    \item w/o AG: removing the correlation-aware encoding.
    \item w/o CDE: removing the continuity-preserving encoding.
\end{itemize}

In each variant, the remaining architecture stays unchanged. Table~\ref{tab:ablation-hr1} shows that the full RePo model achieves the best HR@1 performance on all metrics and datasets, validating the effectiveness of multi-branch fusion and multi-scale feature extraction. Removing the CNN branch causes the largest drop, highlighting the importance of local pattern extraction. Omitting the ResNet or node2vec modules reduces the model's ability to capture spatial and structural grid information, while removing the graph or CDE branches weakens correlation and continuity modeling. These results confirm the complementary strengths of each module and the benefit of joint representation.

\begin{table}[!t]
\centering
\small

\begin{tabular}{l|cc|cc}
\toprule
\multirow{2}{*}{Method} & \multicolumn{2}{c|}{DFD} & \multicolumn{2}{c}{DTW} \\
\cmidrule(lr){2-3}\cmidrule(lr){4-5}
 & Porto & Chengdu & Porto & Chengdu \\
\midrule
w/o ResNet   & 0.432 & 0.332 & 0.491 & 0.369 \\
w/o node2vec & 0.434 & 0.273 & 0.468 & 0.307 \\
w/o CNN      & 0.374 & 0.284 & 0.478 & 0.344 \\
w/o AG       & 0.434 & 0.310 & 0.467 & 0.392 \\
w/o CDE      & 0.425 & 0.372 & 0.462 & 0.403 \\
\textbf{RePo}         & \textbf{0.483} & \textbf{0.392} & \textbf{0.532} & \textbf{0.428} \\
\bottomrule
\end{tabular}

\caption{Ablation study on Porto and Chengdu datasets.}
\label{tab:ablation-hr1}
\end{table}

\stitle{Scalable Experiment.} To evaluate scalability, we directly test the model trained on 10,000 Porto trajectories on larger samples of 100,000 and 200,000 trajectories (Table~\ref{tab:porto-both}). As the test set size increases, the retrieval task becomes more challenging for all methods, leading to an overall decline in performance. This is primarily because the model is trained on a smaller dataset, and the candidate pool becomes significantly larger. Despite this, RePo consistently outperforms NeuTraj and SIMformer across all metrics. This strong performance is attributed to RePo’s multi-modal, discriminative representations, which generalize better to large and diverse candidate sets, confirming its robustness and scalability for real-world trajectory retrieval.

\setlength{\tabcolsep}{0.5mm}
\begin{table}[!t]
\centering
\small

\begin{tabular}{llcccc}
\toprule
\textbf{Datasize} & \textbf{Method} & \textbf{HR@1} & \textbf{R5@20} & \textbf{MRR} & \textbf{NDCG@50} \\
\midrule
\multirow{3}{*}{100k}
    & NeuTraj    & 0.143 & 0.438 & 0.236 & 0.429 \\
    & SIMformer  & 0.123 & 0.427 & 0.213 & 0.433 \\
    & \textbf{RePo}      & \textbf{0.300} & \textbf{0.788} & \textbf{0.448} & \textbf{0.725} \\
\midrule
\multirow{3}{*}{200k}
    & NeuTraj    & 0.111 & 0.350 & 0.188 & 0.351 \\
    & SIMformer  & 0.071 & 0.200 & 0.113 & 0.224 \\
    & \textbf{RePo}      & \textbf{0.192} & \textbf{0.555} & \textbf{0.302} & \textbf{0.508} \\
\bottomrule
\end{tabular}

\caption{
Performance comparison on the 100k and 200k Porto dataset. The best results are highlighted in bold.
}
\label{tab:porto-both}
\end{table}

\begin{figure}[!t]
    \centering
    \includegraphics[width=\columnwidth]{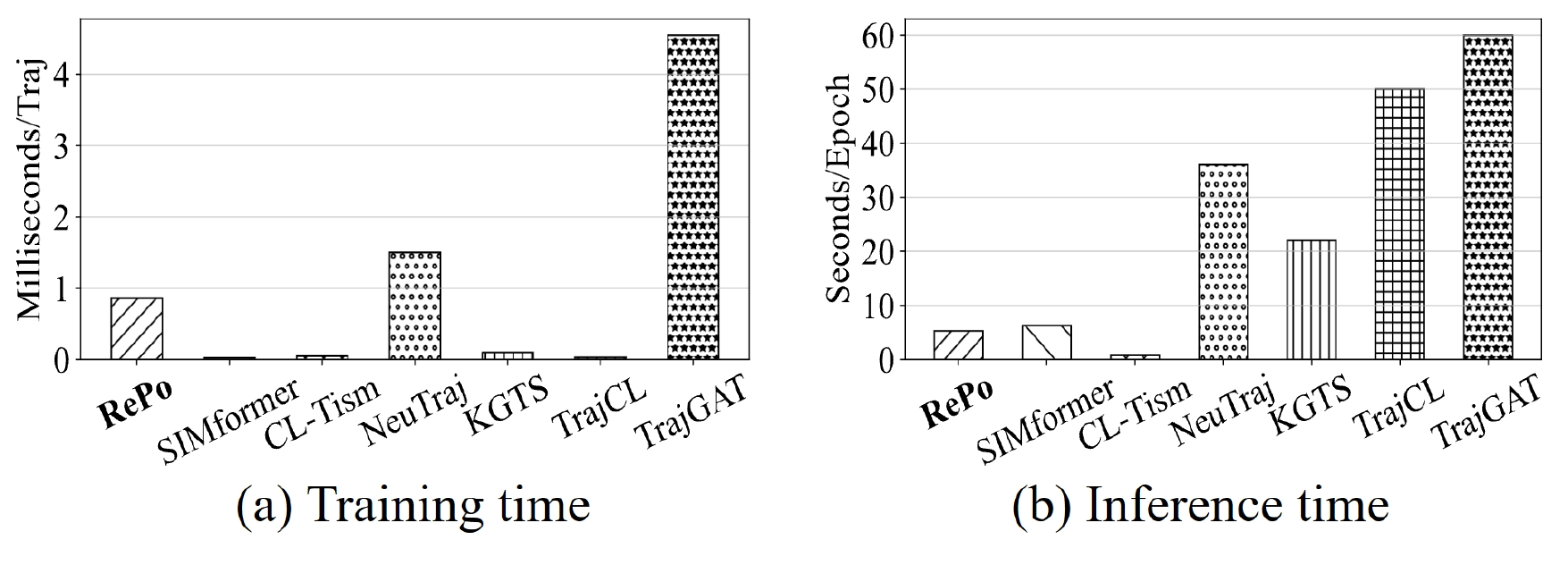}
    \caption{Model training time for one epoch under batch size 128 and model inference time of one trajectory.}
    \label{fig:efficiency}
\end{figure}

\stitle{Efficiency Study.} We compare the training and inference efficiency of different methods on the Porto dataset. The results of Figure~\ref{fig:efficiency} show that RePo achieves significantly faster training than TrajCL and TrajGAT, and maintains moderate inference speed, while consistently delivering the best retrieval accuracy. This demonstrates that RePo offers an effective trade-off between efficiency and performance, making it practical for real-world applications.

\begin{figure}
    \centering
    \includegraphics[width=0.9\columnwidth]{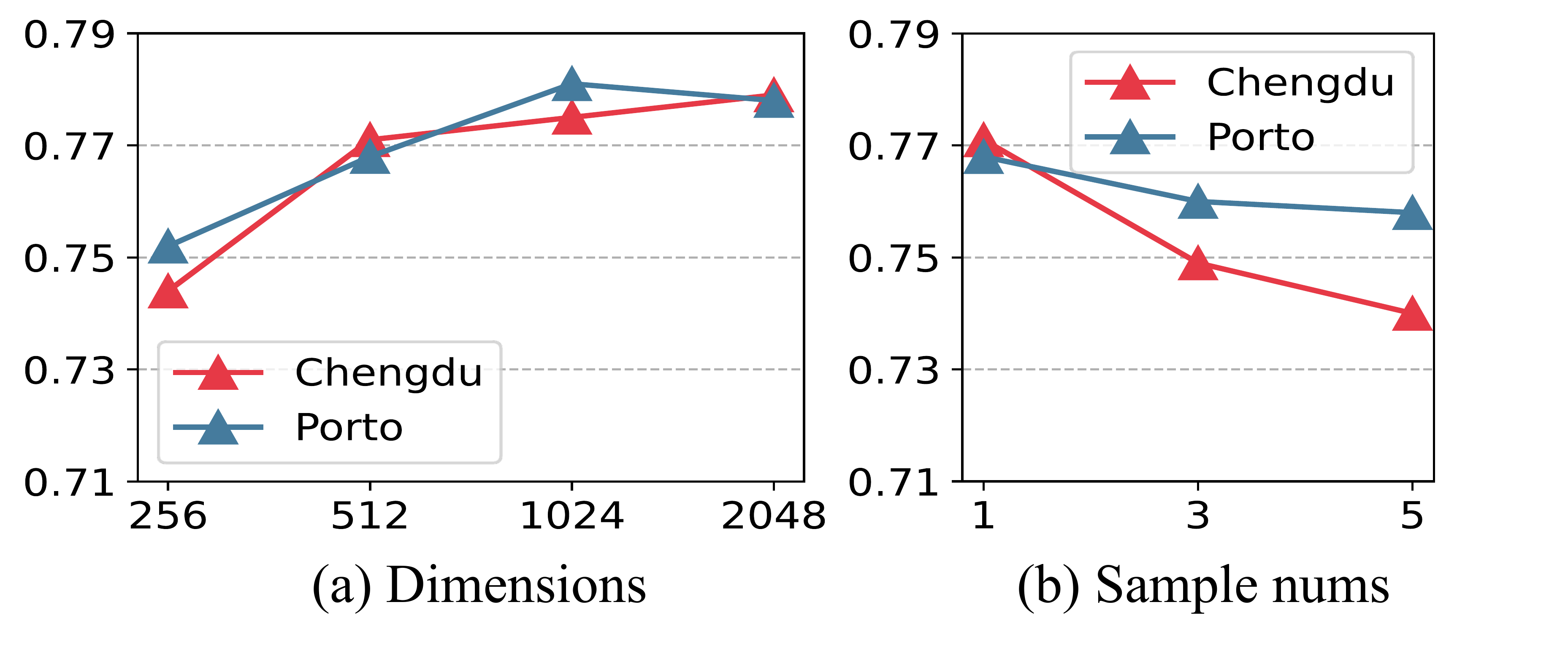}
    \caption{Hyperparameter sensitivity analysis.}
    \label{fig:hyperparam}
\end{figure}

\stitle{Embedding Dimension.} We experiment on how embedding dimension affects RePo's performance. 
As shown in Figure~\ref{fig:hyperparam}(a), increasing the embedding dimension from 256 to 512 leads to a substantial improvement in HR@50, with performance saturating beyond 512. Further increasing the dimension to 1024 or 2048 yields negligible or inconsistent gains, while incurring higher computational and memory costs. We thus select 512 as the default embedding dimension to balance accuracy and efficiency.

\stitle{Number of Hard Negative Samples.}
Figure~\ref{fig:hyperparam}(b) shows the impact of the number of hard negative samples used in the contrastive loss (measured by HR@50). The best performance is obtained with a single hard negative. Using more negatives can introduce noise, especially in the early training stage when representations are not yet stable, which may lead to suboptimal learning. Furthermore, since our goal is to learn correct ranking and the loss primarily enforces pairwise discrimination, a single hard negative is sufficient to drive effective supervision, while additional negatives may dilute the learning signal and even mislead optimization.

\section{Conclusion}
We proposed RePo, a novel multimodal trajectory similarity learning method that integrates region-wise and point-wise information through a combination of structural, visual, and trajectory features. By adopting a mixture-of-experts paradigm and a hierarchical attention-based fusion mechanism, RePo effectively captures the complementary strengths of diverse feature modalities and models fine-grained trajectory characteristics. Extensive experiments on multiple real-world datasets show that RePo consistently achieves better performance compared to SOTA baselines.

\section*{Acknowledgments}
This paper was supported by the NSFC U22B2037 and U21B2046.
\bibliography{RePo}

\section*{Appendix}
\subsection*{Data Preprocessing}
Prior to feature extraction, we apply a preprocessing pipeline to clean the raw GPS trajectory data. Specifically, redundant and outlier points are removed based on a distance-based thresholding criterion: consecutive points separated by a geodesic distance below a small tolerance are considered duplicates and collapsed, while isolated points exhibiting abnormally large displacement relative to their neighbors are treated as outliers and filtered out. The exact implementation follows the heuristic detailed in our open-source code.

During feature extraction, boundary effects arise at the start and end of each trajectory, where the previous point $p_{i-1}$ (for $i=0$) or the next point $p_{i+1}$ (for $i=n_2-1$) does not exist, making the computation of $d_{i-1,i}$, $\theta_{i-1,i}$ or $d_{i,i+1}$, $\theta_{i,i+1}$ undefined. To ensure consistent feature dimensions and smoother representations, we adopt a replication strategy: for the first point, we set $(d_{-1,0}, \theta_{-1,0}) := (d_{0,1}, \theta_{0,1})$, and for the last point, we set $(d_{n_2-1,n_2}, \theta_{n_2-1,n_2}) := (d_{n_2-2,n_2-1}, \theta_{n_2-2,n_2-1})$. This symmetric padding preserves local motion patterns at trajectory boundaries without introducing arbitrary values.

\subsection*{Evaluation Metrics}
\noindent
\textbf{Hit Ratio at $K$ (HR@$K$)} measures the proportion of ground-truth top-$K$ similar trajectories retrieved among the top-$K$ predictions for each query:
\begin{equation}
    \mathrm{HR}@K = \frac{1}{N K} \sum_{i=1}^{N} \left| \mathcal{P}_i^{(K)} \cap \mathcal{G}_i^{(K)} \right|,
\end{equation}
where $\mathcal{P}_i^{(K)}$ and $\mathcal{G}_i^{(K)}$ denote the predicted and ground-truth top-$K$ lists for query $i$, and $N$ is the number of queries.

\stitle{Partial Top-$K$ Hit Ratio (R$m$@$K$)} reports the proportion of ground-truth top-$m$ similar trajectories that are retrieved in the top-$K$ predictions:
\begin{equation}
    \mathrm{R}m@K = \frac{1}{N m} \sum_{i=1}^{N} \left| \mathcal{P}_i^{(K)} \cap \mathcal{G}_i^{(m)} \right|.
\end{equation}
For example, $\mathrm{R}5@20$ measures how many of the ground-truth top-5 are found among the top-20 predictions.

\stitle{Mean Reciprocal Rank (MRR)} quantifies how highly the most relevant trajectory is ranked, and is defined as:
\begin{equation}
    \mathrm{MRR} = \frac{1}{N} \sum_{i=1}^{N} \frac{1}{\mathrm{rank}_i},
\end{equation}
where $\mathrm{rank}_i$ is the rank position of the ground-truth most similar trajectory for query $i$.

\stitle{Normalized Discounted Cumulative Gain (NDCG)} assesses the quality of the top-$K$ ranked list by considering both the relevance and the position of ground-truth trajectories:
\begin{equation}
    \mathrm{NDCG}@K = \frac{1}{N} \sum_{i=1}^{N} \frac{\sum_{j=1}^{K} \frac{\mathrm{rel}_{i,j}}{\log_2(j+1)}}{\sum_{j=1}^{K} \frac{1}{\log_2(j+1)}},
\end{equation}
where $\mathrm{rel}_{i,j}$ is 1 if the $j$-th predicted trajectory is among the ground-truth top-$K$ for query $i$, and 0 otherwise.

\setlength{\tabcolsep}{1mm} 
\begin{table*}[h]
\centering
\small

\scalebox{0.85}{
\begin{tabular}{l|l|cccccccccccc}
\toprule
\multirow{2}{*}{Dataset} 
& \multirow{2}{*}{Method}
& \multicolumn{12}{c}{DFD}
\\
\cmidrule{3-6} \cmidrule{7-10} \cmidrule{11-14}
&
& HR@1 & HR@5 & HR@10 & HR@20
& HR@50 & R5@20 & R10@50 & MRR
& NDCG@5 & NDCG@10 & NDCG@20 & NDCG@50 \\
\midrule
\multirow{9}{*}{Porto}
&t2vec      &0.213 &0.291 &0.314 &0.337 &0.370 &0.556 &0.639 &0.337 &0.323 &0.361 &0.393 &0.433 \\
&CL-TSim    &0.101 &0.136 &0.153 &0.165 &0.184 &0.317 &0.395 &0.184 &0.150 &0.172 &0.190 &0.213 \\
&NeuTraj    &0.372 &0.531 &\underline{0.586} &\underline{0.636} &\underline{0.698} &\underline{0.840} &0.914 &0.536 &0.576 &\underline{0.643} &\underline{0.697} &\underline{0.755} \\
&TrajCL     &0.326 &0.460 &0.510 &0.559 &0.622 &0.803 &0.894 &0.483 &0.508 &0.573 &0.627 &0.687 \\
&TrajGAT    &0.346 &0.412 &0.423 &0.420 &0.429 &0.748 &0.827 &0.497 &0.453 &0.470 &0.475 &0.475 \\
&KGTS       &0.215 &0.263 &0.271 &0.272 &0.273 &0.392 &0.471 &0.321 &0.299 &0.320 &0.328 &0.333 \\
&SIMformer  &\underline{0.422} &\underline{0.535} &0.570 &0.604 &0.653 &0.828 &\underline{0.916} &\underline{0.575} &\underline{0.586} &0.633 &0.670 &0.714 \\
&\textbf{RePo}   &\textbf{0.483} &\textbf{0.614} &\textbf{0.647} &\textbf{0.675} &\textbf{0.708} &\textbf{0.922} &\textbf{0.963} &\textbf{0.640} &\textbf{0.670} &\textbf{0.715} &\textbf{0.745} &\textbf{0.771} \\
\cmidrule{2-14}
&\textbf{Improv.}      &\textbf{14.5\%} &\textbf{14.8\%}  & \textbf{10.4\%} &\textbf{6.13\%}  & \textbf{1.43\%} & \textbf{9.76\%} & \textbf{5.13\%} & \textbf{11.3\%} & \textbf{14.3\%} & \textbf{11.2\%} & \textbf{6.89\%} & \textbf{2.12\%} \\
\bottomrule
\midrule
\multirow{9}{*}{Geolife}
&t2vec      &0.180 &0.214 &0.233 &0.257 &0.291 &0.401 &0.468 &0.260 &0.239 &0.267 &0.297 &0.334 \\
&CL-TSim    &0.166 &0.247 &0.291 &0.336 &0.374 &0.506 &0.608 &0.270 &0.272 &0.327 &0.382 &0.430 \\
&NeuTraj    &0.289 &0.378 &0.428 &0.488 &0.562 &0.713 &\underline{0.825} &0.403 &0.424 &0.489 &0.556 &0.630 \\
&TrajCL     &0.129 &0.226 &0.278 &0.336 &0.408 &0.479 &0.605 &0.238 &0.248 &0.312 &0.379 &0.457 \\
&TrajGAT    &0.283 &0.376 &0.458 &0.426 &0.534 &0.719 &0.821 &0.426 &0.407 &0.465 &0.520 &0.574 \\
&KGTS       &0.206 &0.205 &0.224 &0.248 &0.272 &0.308 &0.396 &0.283 &0.242 &0.271 &0.306 &0.338 \\
&SIMformer  &\underline{0.316} &\underline{0.405} &\underline{0.465} &\underline{0.533} &\underline{0.611} &\underline{0.730} &\underline{0.825} &\underline{0.452} &\underline{0.440} &\underline{0.509} &\underline{0.602} &\underline{0.677} \\
&\textbf{RePo}   &\textbf{0.325} &\textbf{0.409} &\textbf{0.473} &\textbf{0.542} &\textbf{0.622} &\textbf{0.745} &\textbf{0.853} &\textbf{0.455} &\textbf{0.453} &\textbf{0.528} &\textbf{0.610} &\textbf{0.678} \\
\cmidrule{2-14}
&\textbf{Improv.}       &\textbf{2.85\%} &\textbf{0.99\%}  & \textbf{1.72\%} &\textbf{1.69\%}  & \textbf{1.80\%} & \textbf{2.05\%} & \textbf{3.40\%} & \textbf{0.66\%} & \textbf{2.95\%} & \textbf{3.73\%} & \textbf{1.33\%} & \textbf{0.15\%} \\
\bottomrule
\midrule
\multirow{9}{*}{Chengdu}
&t2vec      &0.306 &0.421 &0.447 &0.452 &0.458 &0.737 &0.792 &0.462 &0.469 &0.511 &0.526 &0.533 \\
&CL-TSim    &0.094 &0.164 &0.201 &0.237 &0.278 &0.400 &0.515 &0.190 &0.178 &0.220 &0.263 &0.310 \\
&NeuTraj    &0.316 &0.490 &0.564 &0.640 &\underline{0.727} &0.830 &0.916 &0.478 &0.534 &0.619 &0.696 &\underline{0.777} \\
&TrajCL     &0.313 &0.485 &0.563 &0.630 &0.713 &0.859 &\underline{0.944} &0.482 &0.527 &0.620 &0.691 &0.768 \\
&TrajGAT    &0.338 &0.448 &0.494 &0.475 &0.507 &0.824 &0.902 &0.505 &0.480 &0.514 &0.534 &0.542 \\
&KGTS       &0.179 &0.262 &0.297 &0.317 &0.342 &0.420 &0.539 &0.294 &0.293 &0.341 &0.373 &0.403 \\
&SIMformer  &\underline{0.387} &\underline{0.532} &\underline{0.593} &\underline{0.642} &0.705 &\underline{0.862} &0.941 &\underline{0.546} &\underline{0.582} &\underline{0.654} &\underline{0.707} &0.763 \\
&\textbf{RePo}   &\textbf{0.392} &\textbf{0.549} &\textbf{0.621} &\textbf{0.677} &\textbf{0.740} &\textbf{0.895} &\textbf{0.963} &\textbf{0.552} &\textbf{0.601} &\textbf{0.684} &\textbf{0.743} &\textbf{0.796} \\
\cmidrule{2-14}
&\textbf{Improv.}       &\textbf{1.30\%} &\textbf{3.20\%}  & \textbf{4.72\%} &\textbf{5.45\%}  & \textbf{1.80\%} & \textbf{3.83\%} & \textbf{2.01\%} & \textbf{1.10\%} & \textbf{3.26\%} & \textbf{4.59\%} & \textbf{5.10\%} & \textbf{2.44\%} \\
\bottomrule
\end{tabular}
}
\caption{Performance comparison on the three datasets under DFD trajectory similarity metrics. Best results are bolded, second-best are underlined. “Improv” shows our improvement over the strongest baseline.
}
\label{tab:Frechet-results}
\end{table*}

\setlength{\tabcolsep}{1mm} 
\begin{table*}[ht]
\centering
\small
\scalebox{0.85}{
\begin{tabular}{l|l|cccccccccccc}
\toprule
\multirow{2}{*}{Dataset} 
& \multirow{2}{*}{Method}
& \multicolumn{12}{c}{EDwP}
\\
\cmidrule{3-6} \cmidrule{7-10} \cmidrule{11-14}
&
& HR@1 & HR@5 & HR@10 & HR@20
& HR@50 & R5@20 & R10@50 & MRR
& NDCG@5 & NDCG@10 & NDCG@20 & NDCG@50 \\
\midrule
\multirow{9}{*}{Porto}
&t2vec      &0.235 &0.329 &0.362 &0.391 &0.426 &0.612 &0.703 &0.366 &0.363 &0.413 &0.454 &0.494 \\
&CL-TSim    &0.114 &0.161 &0.183 &0.205 &0.234 &0.357 &0.445 &0.203 &0.179 &0.209 &0.237 &0.273 \\
&NeuTraj    &\underline{0.346} &\underline{0.467} &\underline{0.503} &\underline{0.533} &\underline{0.572} &\underline{0.787} &\underline{0.864} &\underline{0.501} &\underline{0.519} &\underline{0.571} &\underline{0.609} &\underline{0.647} \\
&TrajCL     &0.197 &0.310 &0.348 &0.377 &0.423 &0.586 &0.688 &0.328 &0.346 &0.402 &0.445 &0.493 \\
&TrajGAT    &0.161 &0.217 &0.235 &0.226 &0.250 &0.413 &0.473 &0.261 &0.247 &0.267 &0.283 &0.298 \\
&KGTS       &0.234 &0.293 &0.302 &0.310 &0.316 &0.420 &0.502 &0.345 &0.333 &0.359 &0.378 &0.388 \\
&SIMformer  &0.321 &0.386 &0.397 &0.407 &0.426 &0.579 &0.782 &0.466 &0.428 &0.449 &0.461 &0.475 \\
&\textbf{RePo}   &\textbf{0.512} &\textbf{0.660} &\textbf{0.705} &\textbf{0.748} &\textbf{0.781} &\textbf{0.947} &\textbf{0.977} &\textbf{0.667} &\textbf{0.715} &\textbf{0.768} &\textbf{0.808} &\textbf{0.833} \\
\cmidrule{2-14}
&\textbf{Improv.}       &\textbf{47.9\%} &\textbf{41.3\%}  & \textbf{40.1\%} &\textbf{40.3\%}  & \textbf{36.5\%} & \textbf{20.3\%} & \textbf{13.1\%} & \textbf{33.1\%} & \textbf{37.8\%} & \textbf{34.5\%} & \textbf{32.7\%} & \textbf{28.7\%} \\
\bottomrule
\midrule
\multirow{9}{*}{Geolife}
&t2vec      &0.171 &0.227 &0.254 &0.286 &0.323 &0.429 &0.506 &0.256 &0.252 &0.291 &0.329 &0.368\\
&CL-TSim    &0.173 &0.278 &0.331 &0.390 &0.436 &0.556 &0.665 &0.285 &0.306 &0.374 &0.443 &0.500\\
&NeuTraj    &\underline{0.250} &\underline{0.328} &0.382 &\underline{0.441} &0.510 &\underline{0.657} &\underline{0.779} &\underline{0.363} &\underline{0.370} &\underline{0.438} &\underline{0.508} &\underline{0.581}\\
&TrajCL     &0.062 &0.124 &0.161 &0.210 &0.275 &0.285 &0.400 &0.134 &0.138 &0.184 &0.242 &0.316\\
&TrajGAT    &0.214 &0.280 &\underline{0.384} &0.327 &0.458 &0.569 &0.693 &0.326 &0.308 &0.364 &0.425 &0.501\\
&KGTS       &0.210 &0.218 &0.239 &0.267 &0.298 &0.323 &0.416 &0.289 &0.259 &0.292 &0.330 &0.369\\
&SIMformer  &0.218 &0.316 &0.378 &0.440 &\underline{0.521} &0.570 &0.776 &0.342 &0.342 &0.412 &0.479 &0.560\\
&\textbf{RePo}   &\textbf{0.347} &\textbf{0.448} &\textbf{0.519} &\textbf{0.590} &\textbf{0.670} &\textbf{0.787} &\textbf{0.891} &\textbf{0.483} &\textbf{0.494} &\textbf{0.575} &\textbf{0.652} &\textbf{0.728}\\
\cmidrule{2-14}
&\textbf{Improv.}       &\textbf{38.8\%} &\textbf{36.6\%} &\textbf{35.2\%}  & \textbf{33.8\%} & \textbf{28.6\%} & \textbf{19.8\%} & \textbf{14.4\%} & \textbf{33.1\%} & \textbf{33.5\%} & \textbf{31.3\%} & \textbf{28.3\%}   & \textbf{25.3\%} \\
\bottomrule
\midrule
\multirow{9}{*}{Chengdu}
&t2vec      &\underline{0.324} &\underline{0.439} &0.476 &0.498 &0.525 &\underline{0.768} &0.836 &\underline{0.487} &\underline{0.485} &\underline{0.539} &0.570 &0.596\\
&CL-TSim    &0.103 &0.181 &0.224 &0.271 &0.326 &0.431 &0.560 &0.204 &0.197 &0.248 &0.303 &0.365\\
&NeuTraj    &0.263 &0.404 &0.466 &0.522 &0.590 &0.740 &0.844 &0.409 &0.447 &0.524 &0.589 &0.657\\
&TrajCL     &0.091 &0.185 &0.228 &0.273 &0.329 &0.407 &0.537 &0.184 &0.205 &0.260 &0.313 &0.376\\
&TrajGAT    &0.100 &0.178 &0.243 &0.209 &0.280 &0.388 &0.499 &0.193 &0.198 &0.240 &0.281 &0.321\\
&KGTS       &0.186 &0.281 &0.323 &0.355 &0.393 &0.432 &0.564 &0.297 &0.316 &0.375 &0.419 &0.463\\
&SIMformer  &0.279 &0.421 &\underline{0.479} &\underline{0.529} &\underline{0.600} &0.755 &\underline{0.859} &0.429 &0.463 &0.536 &\underline{0.597} &\underline{0.667}\\
&\textbf{RePo}  &\textbf{0.360} &\textbf{0.556} &\textbf{0.636} &\textbf{0.703} &\textbf{0.779} &\textbf{0.904} &\textbf{0.971} &\textbf{0.526} &\textbf{0.605} &\textbf{0.697} &\textbf{0.765} &\textbf{0.830}\\
\cmidrule{2-14}
&\textbf{Improv.}       &\textbf{11.1\%} &\textbf{26.7\%}  & \textbf{32.8\%} &\textbf{32.9\%}  & \textbf{29.8\%} & \textbf{17.7\%} & \textbf{13.0\%} & \textbf{8.01\%} & \textbf{24.7\%} & \textbf{29.3\%} & \textbf{33.2\%} & \textbf{24.7\%} \\
\bottomrule
\end{tabular}
}
\caption{Performance comparison on the three datasets under EDwP trajectory similarity metrics. Best results are bolded, second-best are underlined. “Improv” shows our improvement over the strongest baseline.
}
\label{tab:EDwP-results}
\end{table*}

\subsection*{Supplementary Experimental Data}

To provide a comprehensive evaluation of model performance, we report a diverse set of retrieval and ranking metrics in addition to the main results. These include Hit Ratio at various cutoffs (HR@1, HR@5, HR@10, HR@20), Partial Hit Ratio (R5@20, R10@50), Mean Reciprocal Rank (MRR), and Normalized Discounted Cumulative Gain (NDCG@5, NDCG@10, NDCG@20, NDCG@50). Reporting results on multiple metrics and cutoff values allows for a more nuanced analysis of both retrieval accuracy and ranking effectiveness under different evaluation criteria.

\begin{itemize}
    \item \textbf{Hit Ratio (HR@K)}: Measures the proportion of queries for which the ground-truth most similar trajectory is present in the top-$K$ retrieved results. Higher HR@K values indicate better retrieval capability, particularly for top-$K$ candidates.
    \item \textbf{Partial Hit Ratio (R$m$@K)}: Assesses the proportion of the ground-truth top-$m$ similar trajectories that are successfully found among the top-$K$ predictions. This metric evaluates the model's ability to retrieve multiple relevant candidates, rather than only the top-1.
    \item \textbf{Mean Reciprocal Rank (MRR)}: Computes the average reciprocal rank of the first relevant item in the retrieved list for each query, reflecting how highly the most similar trajectory is positioned by the model.
    \item \textbf{Normalized Discounted Cumulative Gain (NDCG@K)}: Measures ranking quality by accounting for both the relevance and rank position of each retrieved trajectory within the top-$K$ list. NDCG rewards models that retrieve relevant items early in the ranked list and is widely used to assess overall ranking effectiveness.
\end{itemize}

By evaluating with multiple thresholds ($K$ and $m$), these metrics jointly reflect the practical utility of a model in both retrieval and ranking tasks.  
All results are computed as averages over the entire test set, ensuring robust and fair comparisons between models.  
Such a comprehensive evaluation framework allows us to systematically analyze model strengths and weaknesses across a spectrum of retrieval and ranking scenarios, supporting more reliable conclusions about the generalization ability and real-world applicability of different methods.

\setlength{\tabcolsep}{1mm} 
\begin{table*}[ht]
\centering
\small

\scalebox{0.85}{
\begin{tabular}{l|l|cccccccccccc}
\toprule
\multirow{2}{*}{Dataset} 
& \multirow{2}{*}{Method}
& \multicolumn{12}{c}{DTW}
\\
\cmidrule{3-6} \cmidrule{7-10} \cmidrule{11-14}
&
& HR@1 & HR@5 & HR@10 & HR@20
& HR@50 & R5@20 & R10@50 & MRR
& NDCG@5 & NDCG@10 & NDCG@20 & NDCG@50 \\
\midrule
\multirow{9}{*}{Porto}
&t2vec      &0.239 &0.337 &0.369 &0.393 &0.425 &0.624 &0.715 &0.373 &0.372 &0.417 &0.454 &0.490\\
&CL-TSim    &0.116 &0.162 &0.184 &0.204 &0.228 &0.362 &0.448 &0.204 &0.180 &0.210 &0.236 &0.266\\
&NeuTraj    &\underline{0.308} &\underline{0.437} &\underline{0.489} &\underline{0.538} &\underline{0.597} &\underline{0.807} &\underline{0.892} &\underline{0.448} &\underline{0.490} &\underline{0.560} &\underline{0.617} &\underline{0.672}\\
&TrajCL     &0.217 &0.329 &0.374 &0.416 &0.482 &0.603 &0.713 &0.343 &0.372 &0.435 &0.489 &0.556\\
&TrajGAT    &0.296 &0.362 &0.393 &0.383 &0.401 &0.683 &0.766 &0.437 &0.401 &0.432 &0.445 &0.450\\
&KGTS       &0.253 &0.314 &0.324 &0.329 &0.336 &0.444 &0.525 &0.369 &0.356 &0.385 &0.402 &0.411\\
&SIMformer  &0.300 &0.423 &0.468 &0.516 &0.584 &0.755 &0.845 &0.446 &0.469 &0.531 &0.587 &0.651\\
&\textbf{RePo}   &\textbf{0.532} &\textbf{0.672} &\textbf{0.715} &\textbf{0.747} &\textbf{0.768} &\textbf{0.955} &\textbf{0.983} &\textbf{0.682} &\textbf{0.726} &\textbf{0.777} &\textbf{0.807} &\textbf{0.821}\\
\cmidrule{2-14}
&\textbf{Improv.}       &\textbf{72.7\%} &\textbf{53.8\%}  & \textbf{46.2\%} &\textbf{38.8\%}  & \textbf{28.6\%} & \textbf{18.3\%} & \textbf{10.2\%} & \textbf{52.2\%} & \textbf{48.2\%} & \textbf{38.7\%} & \textbf{30.8\%} & \textbf{22.2\%} \\
\bottomrule
\midrule
\multirow{9}{*}{Geolife}
&t2vec      &0.188 &0.230 &0.256 &0.283 &0.321 &0.430 &0.510 &0.273 &0.256 &0.292 &0.327 &0.368\\
&CL-TSim    &0.181 &\underline{0.288} &\underline{0.343} &\underline{0.396} &0.434 &0.571 &0.677 &0.295 &\underline{0.317} &\underline{0.384} &\underline{0.448} &\underline{0.497}\\
&NeuTraj    &0.202 &0.254 &0.289 &0.328 &0.382 &0.532 &0.648 &0.287 &0.292 &0.342 &0.394 &0.455\\
&TrajCL     &0.107 &0.172 &0.209 &0.251 &0.310 &0.357 &0.469 &0.195 &0.194 &0.242 &0.295 &0.361\\
&TrajGAT    &0.212 &0.274 &0.376 &0.322 &\underline{0.436} &\underline{0.587} &\underline{0.716} &\underline{0.335} &0.300 &0.355 &0.412 &0.477\\
&KGTS       &\underline{0.219} &0.224 &0.244 &0.270 &0.295 &0.324 &0.413 &0.296 &0.264 &0.297 &0.333 &0.366\\
&SIMformer  &0.194 &0.268 &0.317 &0.360 &0.419 &0.525 &0.609 &0.288 &0.298 &0.358 &0.412 &0.474\\
&\textbf{RePo}   &\textbf{0.342} &\textbf{0.433} &\textbf{0.496} &\textbf{0.557} &\textbf{0.628} &\textbf{0.776} &\textbf{0.881} &\textbf{0.475} &\textbf{0.475} &\textbf{0.550} &\textbf{0.617} &\textbf{0.685}\\
\cmidrule{2-14}
&\textbf{Improv.}       &\textbf{56.2\%} &\textbf{50.3\%}  & \textbf{44.6\%} &\textbf{40.6\%}  & \textbf{44.0\%} & \textbf{32.2\%} & \textbf{23.0\%} & \textbf{41.8\%} & \textbf{49.8\%} & \textbf{43.2\%} & \textbf{37.7\%} & \textbf{37.8\%} \\
\bottomrule
\midrule
\multirow{9}{*}{Chengdu}
&t2vec      &\underline{0.330} &\underline{0.448} &\underline{0.488} &\underline{0.512} &0.531 &\underline{0.792} &\underline{0.857} &\underline{0.493} &\underline{0.494} &\underline{0.550} &\underline{0.582}&0.602\\
&CL-TSim    &0.114 &0.193 &0.235 &0.278 &0.326 &0.444 &0.572 &0.215 &0.213 &0.263 &0.313 &0.367\\
&NeuTraj    &0.249 &0.374 &0.421 &0.467 &\underline{0.532} &0.722 &0.833 &0.385 &0.419 &0.487 &0.544 &\underline{0.609}\\
&TrajCL     &0.110 &0.196 &0.229 &0.259 &0.304 &0.409 &0.505 &0.206 &0.218 &0.264 &0.304 &0.354\\
&TrajGAT    &0.190 &0.299 &0.371 &0.341 &0.402 &0.618 &0.732 &0.327 &0.326 &0.379 &0.417 &0.447\\
&KGTS       &0.225 &0.325 &0.363 &0.387 &0.414 &0.491 &0.611 &0.353 &0.364 &0.419 &0.454 &0.485\\
&SIMformer  &0.235 &0.354 &0.404 &0.453 &\underline{0.532} &0.706 &0.744 &0.369 &0.401 &0.471 &0.531 &0.608\\
&\textbf{RePo}   &\textbf{0.428} &\textbf{0.598} &\textbf{0.663} &\textbf{0.720} &\textbf{0.771} &\textbf{0.929} &\textbf{0.979} &\textbf{0.592} &\textbf{0.653} &\textbf{0.728} &\textbf{0.782} &\textbf{0.823}\\
\cmidrule{2-14}
&\textbf{Improv.}       &\textbf{29.7\%} &\textbf{33.5\%}  & \textbf{35.9\%} &\textbf{40.6\%}  & \textbf{44.9\%} & \textbf{17.3\%} & \textbf{14.2\%} & \textbf{20.1\%} & \textbf{32.2\%} & \textbf{32.4\%} & \textbf{34.4\%} & \textbf{35.1\%} \\
\bottomrule
\end{tabular}
}
\caption{Performance comparison on the three datasets under DTW trajectory similarity metrics. Best results are bolded, second-best are underlined. “Improv” shows our improvement over the strongest baseline.
}
\label{tab:DTW-results}
\end{table*}

\subsection*{Results Analysis}
Tables~\ref{tab:Frechet-results}--\ref{tab:DTW-results} report comprehensive performance comparisons of all methods on the Porto, Geolife, and Chengdu datasets under three similarity metrics (Discret Fréchet(DFD), EDwP, and Dynamic Time Wraping(DTW)). We evaluate using a variety of retrieval and ranking metrics, including HR@K, R$m$@K, MRR, and NDCG@K.

Across all datasets and similarity measures, RePo consistently achieves the best results on all metrics, significantly outperforming prior state-of-the-art methods. Notably, RePo yields the largest improvements on the Porto dataset and under the DTW and EDwP metrics, where it surpasses the strongest baseline by up to 72.7\% (HR@1, DTW, Porto). Substantial gains are also observed in MRR and NDCG metrics, indicating both high retrieval accuracy and ranking quality.

On the Geolife and Chengdu datasets, while all methods exhibit lower overall performance (due to increased data sparsity and complexity), RePo maintains clear advantages, especially on ranking-oriented metrics. Under the DFD metric, improvements are relatively less pronounced, as it primarily emphasizes global shape alignment, whereas our model excels at capturing multi-scale and fine-grained spatial patterns.

Overall, these results confirm the effectiveness and robustness of RePo’s multi-modal and multi-scale trajectory representation, as well as its strong generalization ability across diverse evaluation scenarios.

\end{document}